\documentclass[conference]{IEEEtran}
\IEEEoverridecommandlockouts

\usepackage[utf8]{inputenc}
\usepackage[T1]{fontenc}
\usepackage{microtype,newtxtext,newtxmath} 
\usepackage{mathtools,gensymb,eurosym,url,soul} 
\usepackage[font=footnotesize,textfont=footnotesize,labelsep=period]{subcaption}
\usepackage{colortbl}
\usepackage{graphicx}
\usepackage{xcolor}
\usepackage{svg}
\usepackage{url}
\usepackage{acronym}
\usepackage{longtable,booktabs,array,multirow,tabularx}
\usepackage[normalem]{ulem}
\usepackage[switch]{lineno}
\usepackage[backend=biber, sorting=none, style=ieee, citestyle=numeric-comp, giveninits=true, maxnames=5]{biblatex}
\usepackage{hyperref}
\hypersetup{colorlinks=true, breaklinks=true,
  urlcolor=blue, linkcolor=blue,anchorcolor=blue,citecolor=blue,
  hypertexnames=true, final=true,
  pdfpagemode = UseNone,
  pdfauthor = {},
  pdftitle = {},
  pdfsubject = {},
  pdfkeywords = {}
}

\graphicspath{{figures/} {figs/} {./}}
\DeclareGraphicsExtensions{.pdf,.png,.jpg,.mps}

\def\BibTeX{{\rm B\kern-.05em{\sc i\kern-.025em b}\kern-.08em
    T\kern-.1667em\lower.7ex\hbox{E}\kern-.125emX}}

\DeclareSourcemap{
  \maps[datatype=bibtex]{
    \map[overwrite=true]{
      \step[fieldset=url, null]
      \step[fieldset=doi, null]
      \step[fieldset=eprint, null]
      \step[fieldset=month, null]
      \step[fieldset=issn, null]
    }
  }
}

\acrodef{ADC}[ADC]{Analog-to-Digital Converter}
\acrodef{ADEXP}[AdExp-IF]{Adaptive Exponential Integrate-and-Fire}
\acrodef{ADM}[ADM]{Asynchronous Delta Modulator}
\acrodef{AER}[AER]{Address-Event Representation}
\acrodef{AEX}[AEX]{AER EXtension board}
\acrodef{AE}[AE]{Address-Event}
\acrodef{AFE}[AFE]{Analog Front-End}
\acrodef{AFM}[AFM]{Atomic Force Microscope}
\acrodef{AGC}[AGC]{Automatic Gain Control}
\acrodef{AI}[AI]{Artificial Intelligence}
\acrodef{AMDA}[AMDA]{AER Motherboard with D/A converters}
\acrodef{AMPA}[AMPA]{$\alpha$-Amino-3-hydroxy-5-methyl-4-isoxazolepropionic Acid}
\acrodef{ANN}[ANN]{Artificial Neural Network}
\acrodef{API}[API]{Application Programming Interface}
\acrodef{APMOM}[APMOM]{Alternate Polarity Metal On Metal}
\acrodef{ARM}[ARM]{Advanced RISC Machine}
\acrodef{ASIC}[ASIC]{Application Specific Integrated Circuit}
\acrodef{AUC}[AUC]{Area Under the Curve}
\acrodef{BCM}[BMC]{Bienenstock-Cooper-Munro}
\acrodef{BD}[BD]{Bundled Data}
\acrodef{BEOL}[BEOL]{Back-end of Line}
\acrodef{BG}[BG]{Bias Generator}
\acrodef{BMI}[BMI]{Brain-Machine Interface}
\acrodef{BTB}[BTB]{Band-to-Band tunnelling}
\acrodef{BTSP}[BTSP]{Behavioral Time Scale Synaptic Plasticity}
\acrodef{CA}{Cortical Automaton}
\acrodefplural{CA}{Cortical Automata}
\acrodef{CAD}[CAD]{Computer Aided Design}
\acrodef{CAM}{Content Addressable Memory}
\acrodefplural{CAM}{Content Addressable Memories}
\acrodef{CAVIAR}[CAVIAR]{Convolution AER Vision Architecture for Real-Time}
\acrodef{CCN}[CCN]{Cooperative and Competitive Network}
\acrodef{CDR}[CDR]{Clock-Data Recovery}
\acrodef{CFC}[CFC]{Current to Frequency Converter}
\acrodef{CHP}[CHP]{Communicating Hardware Processes}
\acrodef{CMIM}[CMIM]{Metal-Insulator-Metal Capacitor}
\acrodef{CML}[CML]{Current Mode Logic}
\acrodef{CMOL}[CMOL]{Hybrid CMOS nanoelectronic circuits}
\acrodef{CMOS}[CMOS]{Complementary Metal-Oxide-Semiconductor}
\acrodef{CNN}[CNN]{Convolutional Neural Network}
\acrodef{CNS}[CNS]{Central Nervous System}
\acrodef{COTS}[COTS]{Commercial Off-The-Shelf}
\acrodef{CPG}[CPG]{Central Pattern Generator}
\acrodef{CPLD}[CPLD]{Complex Programmable Logic Device}
\acrodef{CPU}[CPU]{Central Processing Unit}
\acrodef{CSM}[CSM]{Cortical State Machine}
\acrodef{CSP}[CSP]{Constraint Satisfaction Problem}
\acrodef{CTXCTL}[CTXCTL]{CortexControl}
\acrodef{CV}{Coefficient of Variation}
\acrodefplural{CV}{Coefficients of Variation}
\acrodef{DAC}[DAC]{Digital to Analog Converter}
\acrodef{DAS}[DAS]{Dynamic Auditory Sensor}
\acrodef{DAVIS}[DAVIS]{Dynamic and Active Pixel Vision Sensor}
\acrodef{DBN}[DBN]{Deep Belief Network}
\acrodef{DBS}[DBS]{Deep Brain Stimulation}
\acrodef{DFA}{Deterministic Finite Automaton}
\acrodefplural{DFA}{Deterministic Finite Automata}
\acrodef{DIBL}[DIBL]{Drain-Induced Barrier-Lowering}
\acrodef{DI}[DI]{Delay Insensitive}
\acrodef{divmod3}[DIVMOD3]{Divisibility of a number by three}
\acrodef{DMA}[DMA]{Direct Memory Access}
\acrodef{DNF}[DNF]{Dynamic Neural Field}
\acrodef{DNN}[DNN]{Deep Neural Network}
\acrodef{DoF}{Degree of Freedom}
\acrodefplural{DoF}{Degrees of Freedom}
\acrodef{DPE}[DPE]{Dynamic Parameter Estimation}
\acrodef{DPI}[DPI]{Differential Pair Integrator}
\acrodef{DRAM}[DRAM]{Dynamic Random Access Memory}
\acrodef{DR}[DR]{Dual Rail}
\acrodef{DRRZ}[DR-RZ]{Dual-Rail Return-to-Zero}
\acrodef{DSP}[DSP]{Digital Signal Processor}
\acrodef{DTP}[DTP]{Difference Target Propagation}
\acrodef{DVS}[DVS]{Dynamic Vision Sensor}
\acrodef{DYNAP}[DYNAP]{Dynamic Neuromorphic Asynchronous Processor}
\acrodef{EBL}[EBL]{Electron Beam Lithography}
\acrodef{ECG}[ECG]{Electrocardiography}
\acrodef{ECoG}[ECoG]{Electrocorticography}
\acrodef{EDVAC}[EDVAC]{Electronic Discrete Variable Automatic Computer}
\acrodef{EEG}[EEG]{Electroencephalography}
\acrodef{EI}[EI]{Excitatory-Inhibitory}
\acrodef{EIN}[EIN]{Excitatory-Inhibitory Network}
\acrodef{EM}[EM]{Expectation Maximization}
\acrodef{EMG}[EMG]{Electromyography}
\acrodef{EOG}[EOG]{Electrooculogram}
\acrodef{EPSC}[EPSC]{Excitatory Post-Synaptic Current}
\acrodef{EPSP}[EPSP]{Excitatory Post-Synaptic Potential}
\acrodef{EZ}[EZ]{Epileptogenic Zone}
\acrodef{FDSOI}[FDSOI]{Fully-Depleted Silicon on Insulator}
\acrodef{FET}[FET]{Field-Effect Transistor}
\acrodef{FFT}[FFT]{Fast Fourier Transform}
\acrodef{FI}[F-I]{Frequency--Current}
\acrodef{FMA}[FMA]{Floating Microelectrode Array}
\acrodef{FNN}[FNN]{Feed-forward Neural Network}
\acrodef{FPGA}[FPGA]{Field Programmable Gate Array}
\acrodef{FR}[FR]{Fast Ripple}
\acrodef{FSA}{Finite State Automaton}
\acrodefplural{FSA}{Finite State Automata}
\acrodef{FSM}[FSM]{Finite State Machine}
\acrodef{GABA}[GABA]{$\gamma$-Aminobutanoic Acid}
\acrodef{GIDL}[GIDL]{Gate-Induced Drain Leakage}
\acrodef{GOPS}[GOPS]{Giga-Operations per Second}
\acrodef{GPIO}[GPIO]{General Purpose I/O}
\acrodef{GPU}[GPU]{Graphical Processing Unit}
\acrodef{GRU}[GRU]{Gated Recurrent Unit}
\acrodef{GT}[GT]{Ground Truth}
\acrodef{GUI}[GUI]{Graphical User Interface}
\acrodef{HAL}[HAL]{Hardware Abstraction Layer}
\acrodef{HFO}[HFO]{High Frequency Oscillation}
\acrodef{HH}[H\&H]{Hodgkin \& Huxley}
\acrodef{HMI}[HMI]{Human Machine Interface}
\acrodef{HMM}[HMM]{Hidden Markov Model}
\acrodef{HR}[HR]{Heart Rate}
\acrodef{HRS}[HRS]{High-Resistive State}
\acrodef{HSE}[HSE]{Handshaking Expansion}
\acrodef{HW}[HW]{Hardware}
\acrodef{hWTA}[hWTA]{Hard Winner-Take-All}
\acrodef{IC}[IC]{Integrated Circuit}
\acrodef{ICA}[ICA]{Independent Component Analysis}
\acrodef{ICT}[ICT]{Information and Communication Technology}
\acrodef{iEEG}[iEEG]{Intracranial Electroencephalography}
\acrodef{IF2DWTA}[IF2DWTA]{Integrate \& Fire 2-Dimensional WTA}
\acrodef{IF}[I\&F]{Integrate-and-Fire}
\acrodef{IFSLWTA}[IFSLWTA]{Integrate \& Fire Stop Learning WTA}
\acrodef{IMU}[IMU]{Inertial Measurement Unit}
\acrodef{INCF}[INCF]{International Neuroinformatics Coordinating Facility}
\acrodef{INI}[INI]{Institute of Neuroinformatics}
\acrodef{IO}[I/O]{Input/Output}
\acrodef{IoT}[IoT]{Internet of Things}
\acrodef{IP}[IP]{Intellectual Property}
\acrodef{IPSC}[IPSC]{Inhibitory Post-Synaptic Current}
\acrodef{IPSP}[IPSP]{Inhibitory Post-Synaptic Potential}
\acrodef{ISI}[ISI]{Inter-Spike Interval}
\acrodef{JFLAP}[JFLAP]{Java - Formal Languages and Automata Package}
\acrodef{LEDR}[LEDR]{Level-Encoded Dual-Rail}
\acrodef{LFP}[LFP]{Local Field Potential}
\acrodef{LIFE}[LIFE]{Longitudinal Intrafascicular Electrodes}
\acrodef{LIF}[LI\&F]{Leaky Integrate-and-Fire}
\acrodef{LLC}[LLC]{Low Leakage Cell}
\acrodef{LMS}[LMS]{Least Mean Squares}
\acrodef{LNA}[LNA]{Low-Noise Amplifier}
\acrodef{LOSO}[LOSO]{Leave One Subject Out}
\acrodef{LPF}[LPF]{Low Pass Filter}
\acrodef{LRS}[LRS]{Low-Resistive State}
\acrodef{LSM}[LSM]{Liquid State Machine}
\acrodef{LTD}[LTD]{Long Term Depression}
\acrodef{LTI}[LTI]{Linear Time-Invariant}
\acrodef{LTP}[LTP]{Long Term Potentiation}
\acrodef{LTU}[LTU]{Linear Threshold Unit}
\acrodef{LUT}[LUT]{Look-Up Table}
\acrodef{LVDS}[LVDS]{Low Voltage Differential Signaling}
\acrodef{MAE}[MAE]{Mean Absolute Error}
\acrodef{MCMC}[MCMC]{Markov-Chain Monte Carlo}
\acrodef{MEA}[MEA]{Multielectrode Arrays}
\acrodef{MEMS}[MEMS]{Micro-electromechanical System}
\acrodef{MFR}[MFR]{Mean Firing Rate}
\acrodef{MIM}[MIM]{Metal Insulator Metal}
\acrodef{MI}[MI]{Mutual Information}
\acrodef{ML}[ML]{Machine Learning}
\acrodef{MLP}[MLP]{Multilayer Perceptron}
\acrodef{MOSCAP}[MOSCAP]{Metal Oxide Semiconductor Capacitor}
\acrodef{MOSFET}[MOSFET]{Metal Oxide Semiconductor Field-Effect Transistor}
\acrodef{MOS}[MOS]{Metal Oxide Semiconductor}
\acrodef{MRI}[MRI]{Magnetic Resonance Imaging}
\acrodef{MVC}[MVC]{Maximum Voluntary Contraction}
\acrodef{NCS}[NCS]{Neuromorphic Cognitive Systems}
\acrodef{NDFSM}[NDFSM]{Non-deterministic Finite State Machine}
\acrodef{ND}[ND]{Noise-Driven}
\acrodef{NEF}[NEF]{Neural Engineering Framework}
\acrodef{NHML}[NHML]{Neuromorphic Hardware Mark-up Language}
\acrodef{NIL}[NIL]{Nano-Imprint Lithography}
\acrodef{NI}[NI]{Neural Interface}
\acrodef{NMDA}[NMDA]{\textit{N}-Methyl-\textsc{d}-aspartate}
\acrodef{NME}[NE]{Neuromorphic Engineering}
\acrodef{NMF}[NMF]{Non-negative Matrix Factorization}
\acrodef{NN}[NN]{Neural Network}
\acrodef{NOC}[NoC]{Network-on-Chip}
\acrodef{NRZ}[NRZ]{Non-Return-to-Zero}
\acrodef{NSM}[NSM]{Neural State Machine}
\acrodef{OLS}[OLS]{Ordinary Leaste Squares}
\acrodef{OR}[OR]{Operating Room}
\acrodef{OTA}[OTA]{Operational Transconductance Amplifier}
\acrodef{PCA}{Principal Component Analysis}
\acrodefplural{PCA}{Principal Component Analyses}
\acrodef{PCB}[PCB]{Printed Circuit Board}
\acrodef{PCHB}[PCHB]{Pre-Charge Half-Buffer}
\acrodef{PCM}{Phase Change Memory}
\acrodefplural{PCM}{Phase Change Memories}
\acrodef{PC}[PC]{Personal Computer}
\acrodef{PDK}[PDK]{Process Design Kit}
\acrodef{PE}[PE]{Phase Encoding}
\acrodef{PFA}{Probabilistic Finite Automaton}
\acrodefplural{PFA}{Probabilistic Finite Automata}
\acrodef{PFC}{Prefrontal Cortex}
\acrodefplural{PFC}{Prefrontal Cortices}
\acrodef{PFM}[PFM]{Pulse Frequency Modulation}
\acrodef{PGA}[PGA]{Programmable Gain Amplifier}
\acrodef{PNI}[PNI]{Peripheral Nerve Interface}
\acrodef{PNS}[PNS]{Peripheral Nervous System}
\acrodef{PPG}[PPG]{Photoplethysmography}
\acrodef{PR}[PR]{Production Rule}
\acrodef{PSC}[PSC]{Post-Synaptic Current}
\acrodef{PSP}[PSP]{Post-Synaptic Potential}
\acrodef{PSTH}[PSTH]{Peri-Stimulus Time Histogram}
\acrodef{PV}[PV]{Parvalbumin}
\acrodef{PYR}[PYR]{Pyramidal}
\acrodef{QDI}[QDI]{Quasi Delay Insensitive}
\acrodef{R}[R]{Ripple}
\acrodef{RA}[RA]{Resected Area}
\acrodef{RAM}{Random Access Memory}
\acrodefplural{RAM}{Random Access Memories}
\acrodef{RDF}[RDF]{Random Dopant Fluctuation}
\acrodef{RELU}[ReLu]{Rectified Linear Unit}
\acrodef{RISC}[RISC]{Reduced Instruction Set Computer}
\acrodef{RLS}[RLS]{Recursive Least-Squares}
\acrodef{RMSE}[RMSE]{Root Mean Square-Error}
\acrodef{RRMSE}[RRMSE]{Relative Root Mean Square Error}
\acrodef{RMS}[RMS]{Root Mean Square}
\acrodef{RNN}[RNN]{Recurrent Neural Network}
\acrodef{ROC}[ROC]{Receiver Operating Characteristic}
\acrodef{ROLLS}[ROLLS]{Reconfigurable On-Line Learning Spiking}
\acrodef{RRAM}[R-RAM]{Resistive Random Access Memory}
\acrodef{RSA}[RSA]{Respiratory Sinus Arrhythmia}
\acrodef{rSNN}[rSNN]{recurrent Spiking Neural Network}
\acrodef{SAC}[SAC]{Selective Attention Chip}
\acrodef{SAE}[SAE]{Spiking Autoencoder}
\acrodef{SAT}[SAT]{Boolean Satisfiability Problem}
\acrodef{SCI}{Spinal Cord Injury}
\acrodefplural{SCI}{Spinal Cord Injuries}
\acrodef{SCX}[SCX]{Silicon CorteX}
\acrodef{SD}[SD]{Signal-Driven}
\acrodef{SEM}[SEM]{Spike-based Expectation Maximization}
\acrodef{SHD}[SHD]{Spiking Heidelberg Digits}
\acrodef{SimCo}[SimCo]{HIT Simultaneous Control}
\acrodef{SLAM}[SLAM]{Simultaneous Localization and Mapping}
\acrodef{SNN}[SNN]{Spiking Neural Network}
\acrodef{SNR}[SNR]{Signal to Noise Ratio}
\acrodef{SOC}[SoC]{System-On-Chip}
\acrodef{SOI}[SOI]{Silicon on Insulator}
\acrodef{SOZ}[SOZ]{Seizure Onset Zone}
\acrodef{SP}{Separation Property}
\acrodefplural{SP}{Separation Properties}
\acrodef{SPI}[SPI]{Serial Peripheral Interface}
\acrodef{SRAM}{Static Random Access Memory}
\acrodefplural{SRAM}{Static Random Access Memory}
\acrodef{SST}[SST]{Somatostatin}
\acrodef{STDP}[STDP]{Spike-Timing Dependent Plasticity}
\acrodef{STD}[STD]{Short-Term Depression}
\acrodef{STP}[STP]{Short-Term Plasticity}
\acrodef{STT-MRAM}{Spin-Transfer Torque Magnetic Random Access Memory}
\acrodefplural{STT-MRAM}{Spin-Transfer Torque Magnetic Random Access Memories}
\acrodef{STT}[STT]{Spin-Transfer Torque}
\acrodef{SVM}[SVM]{Support Vector Machine}
\acrodef{SW}[SW]{Software}
\acrodef{sWTA}[sWTA]{soft Winner-Take-All}
\acrodef{TCAM}{Ternary Content-Addressable Memory}
\acrodefplural{TCAM}{Ternary Content-Addressable Memories}
\acrodef{TFT}[TFT]{Thin Film Transistor}
\acrodef{TIME}[TIME]{Transverse Intrafascicular Multichannel Electrode}
\acrodef{TLE}{Temporal Lobe Epilepsy}
\acrodefplural{TLE}{Temporal Lobe Epilepsies}
\acrodef{UEA}[UEA]{Utah Electrode Array}
\acrodef{USB}{Universal Serial Bus}
\acrodefplural{USB}{Universal Serial Buses}
\acrodef{USEA}[USEA]{Utah Slanted Electrode Array}
\acrodef{VHDL}[VHDL]{VHSIC Hardware Description Language}
\acrodef{VHSIC}[VHSIC]{Very High Speed Integrated Circuits}
\acrodef{VIP}[VIP]{Vasoactive Intestinal Peptide}
\acrodef{VLSI}[VLSI]{Very Large Scale Integration}
\acrodef{VNS}[VNS]{Vagal Nerve Stimulation}
\acrodef{VOR}{Vestibulo-Ocular Reflex}
\acrodefplural{VOR}{Vestibulo-Ocular Reflexes}
\acrodef{VSA}[VSA]{Vector Symbolic Architecture}
\acrodef{WCST}[WCST]{Wisconsin Card Sorting Test}
\acrodef{WTA}[WTA]{Winner-Take-All}
\acrodef{XML}[XML]{eXtensible Mark-up Language}

\DeclareUnicodeCharacter{00E9}{\'e}

\begin{document}

\acrodef{ADC}[ADC]{Analog to Digital Converter}
\acrodef{ADEXP}[AdExp-I\&F]{Adaptive-Exponential Integrate and Fire}
\acrodef{AdExp}[AdExp-IF]{Adaptive Exponential Integrate-and-Fire}
\acrodef{AE}[AE]{Address-Event}
\acrodef{AER}[AER]{Address-Event Representation}
\acrodef{AEX}[AEX]{AER EXtension board}
\acrodef{AFM}[AFM]{Atomic Force Microscope}
\acrodef{AGC}[AGC]{Automatic Gain Control}
\acrodef{AI}[AI]{Artificial Intelligence}
\acrodef{AMDA}[AMDA]{AER Motherboard with D/A converters}
\acrodef{ANN}[ANN]{Artificial Neural Network}
\acrodef{API}[API]{Application Programming Interface}
\acrodef{APMOM}[APMOM]{Alternate Polarity Metal On Metal}
\acrodef{ARM}[ARM]{Advanced RISC Machine}
\acrodef{ASIC}[ASIC]{Application Specific Integrated Circuit}
\acrodef{BCM}[BMC]{Bienenstock-Cooper-Munro}
\acrodef{BD}[BD]{Bundled Data}
\acrodef{BEOL}[BEOL]{Back-end of Line}
\acrodef{BG}[BG]{Bias Generator}
\acrodef{BL}[BL]{Bit Line}
\acrodef{BMI}[BMI]{Brain-Machince Interface}
\acrodef{BPTT}[BPTT]{Backpropagation Through Time}
\acrodef{BTB}[BTB]{band-to-band tunnelling}
\acrodef{CA}[CA]{Cortical Automaton}
\acrodef{CAD}[CAD]{Computer Aided Design}
\acrodef{CAM}[CAM]{Content Addressable Memory}
\acrodef{CAVIAR}[CAVIAR]{Convolution AER Vision Architecture for Real-Time}
\acrodef{CCN}[CCN]{Cooperative and Competitive Network}
\acrodef{CDR}[CDR]{Clock-Data Recovery}
\acrodef{CFC}[CFC]{Current to Frequency Converter}
\acrodef{CHP}[CHP]{Communicating Hardware Processes}
\acrodef{CIM}[CIM]{Compute In Memory}
\acrodef{CMIM}[CMIM]{Metal-insulator-metal Capacitor}
\acrodef{CML}[CML]{Current Mode Logic}
\acrodef{CMOL}[CMOL]{Hybrid CMOS nanoelectronic circuits}
\acrodef{CMOS}[CMOS]{Complementary Metal-Oxide-Semiconductor}
\acrodef{CNN}[CNN]{Convolutional Neural Network}
\acrodef{COTS}[COTS]{Commercial Off-The-Shelf}
\acrodef{CPG}[CPG]{Central Pattern Generator}
\acrodef{CPLD}[CPLD]{Complex Programmable Logic Device}
\acrodef{CPU}[CPU]{Central Processing Unit}
\acrodef{CSM}[CSM]{Cortical State Machine}
\acrodef{CSP}[CSP]{Constraint Satisfaction Problem}
\acrodef{CTXCTL}[CTXCTL]{CortexControl}
\acrodef{CV}[CV]{Coefficient of Variation}
\acrodef{DAC}[DAC]{Digital to Analog Converter}
\acrodef{DAS}[DAS]{Dynamic Auditory Sensor}
\acrodef{DAVIS}[DAVIS]{Dynamic and Active Pixel Vision Sensor}
\acrodef{DBN}[DBN]{Deep Belief Network}
\acrodef{DFA}[DFA]{Deterministic Finite Automaton}
\acrodef{DI}[DI]{delay insensitive}
\acrodef{DIBL}[DIBL]{drain-induced-barrier-lowering}
\acrodef{divmod3}[DIVMOD3]{divisibility of a number by three}
\acrodef{DMA}[DMA]{Direct Memory Access}
\acrodef{DNF}[DNF]{Dynamic Neural Field}
\acrodef{DNN}[DNN]{Deep Neural Network}
\acrodef{DOF}[DOF]{Degrees of Freedom}
\acrodef{DPE}[DPE]{Dynamic Parameter Estimation}
\acrodef{DPI}[DPI]{Differential Pair Integrator}
\acrodef{DR}[DR]{Dual Rail}
\acrodef{DRAM}[DRAM]{Dynamic Random Access Memory}
\acrodef{DRRZ}[DR-RZ]{Dual-Rail Return-to-Zero}
\acrodef{DSP}[DSP]{Digital Signal Processor}
\acrodef{DVS}[DVS]{Dynamic Vision Sensor}
\acrodef{DYNAP}[DYNAP]{Dynamic Neuromorphic Asynchronous Processor}
\acrodef{EBL}[EBL]{Electron Beam Lithography}
\acrodef{ECG}[ECG]{Electrocardiography}
\acrodef{EDVAC}[EDVAC]{Electronic Discrete Variable Automatic Computer}
\acrodef{EEG}[EEG]{Electroencephalography}
\acrodef{EIN}[EIN]{Excitatory-Inhibitory Network}
\acrodef{EM}[EM]{Expectation Maximization}
\acrodef{EMG}[EMG]{Electromyography}
\acrodef{eNVM}[eNMV]{Embedded Non-Volatile Memory}
\acrodef{EPSC}[EPSC]{Excitatory Post-Synaptic Current}
\acrodef{EPSP}[EPSP]{Excitatory Post-Synaptic Potential}
\acrodef{ES}[ES]{Evolutionary Strategies}
\acrodef{ESN}[ESN]{Echo state Network }
\acrodef{EZ}[EZ]{Epileptogenic Zone}
\acrodef{FDSOI}[FDSOI]{Fully-Depleted Silicon on Insulator}
\acrodef{FET}[FET]{Field-Effect Transistor}
\acrodef{FFT}[FFT]{Fast Fourier Transform}
\acrodef{FI}[F-I]{Frequency-Current}
\acrodef{fMRI}[fMRI]{functional Magnetic Resonance Imaging}
\acrodef{FPGA}[FPGA]{Field Programmable Gate Array}
\acrodef{FR}[FR]{Fast Ripple}
\acrodef{FSA}[FSA]{Finite State Automaton}
\acrodef{FSM}[FSM]{Finite State Machine}
\acrodef{GIDL}[GIDL]{gate-induced-drain-leakage}
\acrodef{GOPS}[GOPS]{Giga-Operations per Second}
\acrodef{GPU}[GPU]{Graphical Processing Unit}
\acrodef{GUI}[GUI]{Graphical User Interface}
\acrodef{HAL}[HAL]{Hardware Abstraction Layer}
\acrodef{HCS}[HCS]{High-Conductive State}
\acrodef{HFO}[HFO]{High Frequency Oscillation}
\acrodef{HH}[H\&H]{Hodgkin \& Huxley}
\acrodef{HMM}[HMM]{Hidden Markov Model}
\acrodef{HR}[HR]{Human Readable}
\acrodef{HRS}[HRS]{High-Resistive State}
\acrodef{HSE}[HSE]{Handshaking Expansion}
\acrodef{HW}[HW]{Hardware}
\acrodef{hWTA}[hWTA]{hard Winner-Take-All}
\acrodef{IC}[IC]{Integrated Circuit}
\acrodef{ICT}[ICT]{Information and Communication Technology}
\acrodef{IEEG}[iEEG]{intracranial electroencephalography}
\acrodef{IF}[I\&F]{Integrate-and-Fire}
\acrodef{IF2DWTA}[IF2DWTA]{Integrate \& Fire 2--Dimensional WTA}
\acrodef{IFSLWTA}[IFSLWTA]{Integrate \& Fire Stop Learning WTA}
\acrodef{IMC}[IMC]{In-Memory Computing}
\acrodef{IMU}[IMU]{Inertial Measurement Unit}
\acrodef{INCF}[INCF]{International Neuroinformatics Coordinating Facility}
\acrodef{INI}[INI]{Institute of Neuroinformatics}
\acrodef{IO}[I/O]{Input/Output}
\acrodef{IoT}[IoT]{Internet of Things}
\acrodef{IP}[IP]{Intellectual Property}
\acrodef{IPSC}[IPSC]{Inhibitory Post-Synaptic Current}
\acrodef{IPSP}[IPSP]{Inhibitory Post-Synaptic Potential}
\acrodef{ISI}[ISI]{Inter-Spike Interval}
\acrodef{JFLAP}[JFLAP]{Java - Formal Languages and Automata Package}
\acrodef{LCS}[LCS]{Low-Conductive State}
\acrodef{LEDR}[LEDR]{Level-Encoded Dual-Rail}
\acrodef{LFP}[LFP]{Local Field Potential}
\acrodef{LIF}[LIF]{Leaky Integrate and Fire}
\acrodef{LLC}[LLC]{Low Leakage Cell}
\acrodef{LLM}[LLM]{Large Language Model}
\acrodef{LNA}[LNA]{Low-Noise Amplifier}
\acrodef{LPF}[LPF]{Low Pass Filter}
\acrodef{LRS}[LRS]{Low-Resistive State}
\acrodef{LSM}[LSM]{Liquid State Machine}
\acrodef{LTD}[LTD]{Long Term Depression}
\acrodef{LTI}[LTI]{Linear Time-Invariant}
\acrodef{LTP}[LTP]{Long Term Potentiation}
\acrodef{LTU}[LTU]{Linear Threshold Unit}
\acrodef{LUT}[LUT]{Look-Up Table}
\acrodef{LVDS}[LVDS]{Low Voltage Differential Signaling}
\acrodef{ML}[ML]{Machine Learning}
\acrodef{MCMC}[MCMC]{Markov-Chain Monte Carlo}
\acrodef{MEMS}[MEMS]{Micro Electro Mechanical System}
\acrodef{MFR}[MFR]{Mean Firing Rate}
\acrodef{MIM}[MIM]{Metal Insulator Metal}
\acrodef{MLP}[MLP]{Multilayer Perceptron}
\acrodef{MOS}[MOS]{Metal Oxide Semiconductor}
\acrodef{MOSCAP}[MOSCAP]{Metal Oxide Semiconductor Capacitor}
\acrodef{MOSFET}[MOSFET]{Metal Oxide Semiconductor Field-Effect Transistor}
\acrodef{MRI}[MRI]{Magnetic Resonance Imaging}
\acrodef{ND}[ND]{Noise-Driven}
\acrodef{NDFSM}[NDFSM]{Non-deterministic Finite State Machine} 
\acrodef{NEF}[NEF]{Neural Engineering Framework}
\acrodef{NHML}[NHML]{Neuromorphic Hardware Mark-up Language}
\acrodef{NIL}[NIL]{Nano-Imprint Lithography}
\acrodef{NMDA}[NMDA]{N-Methyl-D-Aspartate}
\acrodef{NME}[NE]{Neuromorphic Engineering}
\acrodef{NN}[NN]{Neural Network}
\acrodef{NRZ}[NRZ]{Non-Return-to-Zero}
\acrodef{NSM}[NSM]{Neural State Machine}
\acrodef{OR}[OR]{Operating Room}
\acrodef{OTA}[OTA]{Operational Transconductance Amplifier}
\acrodef{PCB}[PCB]{Printed Circuit Board}
\acrodef{PCHB}[PCHB]{Pre-Charge Half-Buffer}
\acrodef{PCM}[PCM]{Phase Change Memory}
\acrodef{PE}[PE]{Processing Element}
\acrodef{PFA}[PFA]{Probabilistic Finite Automaton}
\acrodef{PFC}[PFC]{prefrontal cortex}
\acrodef{PFM}[PFM]{Pulse Frequency Modulation}
\acrodef{PR}[PR]{Production Rule}
\acrodef{PSC}[PSC]{Post-Synaptic Current}
\acrodef{PSP}[PSP]{Post-Synaptic Potential}
\acrodef{PSTH}[PSTH]{Peri-Stimulus Time Histogram}
\acrodef{PVT}[PVT]{Process, Voltage, Temprature}
\acrodef{QDI}[QDI]{Quasi Delay Insensitive}
\acrodef{R}[R]{Ripples}
\acrodef{RAM}[RAM]{Random Access Memory}
\acrodef{RDF}[RDF]{random dopant fluctuation}
\acrodef{RELU}[ReLu]{Rectified Linear Unit}
\acrodef{RL}[RL]{Reinforcement Learning}
\acrodef{RLS}[RLS]{Recursive Least-Squares}
\acrodef{RMS}[RMS]{Root Mean Squared}
\acrodef{RMSE}[RMSE]{Root Mean Squared-Error}
\acrodef{MSE}[MSE]{Mean Squared-Error}
\acrodef{KL}[KL]{Kullback-Leibler}
\acrodef{RNN}[RNN]{Recurrent Neural Networks}
\acrodef{ROLLS}[ROLLS]{Reconfigurable On-Line Learning Spiking}
\acrodef{RRAM}[RRAM]{Resistive Random Access Memory}
\acrodef{RSNN}[RSNN]{Recurrent Spiking Neural Network}
\acrodef{SAT}[SAT]{Boolean Satisfiability Problem}
\acrodef{SCX}[SCX]{Silicon CorteX}
\acrodef{SD}[SD]{Signal-Driven}
\acrodef{SEM}[SEM]{Spike-based Expectation Maximization}
\acrodef{SHD}[SHD]{Spiking Heidelberg Digit}
\acrodef{SL}[SL]{Source Line}
\acrodef{SLAM}[SLAM]{Simultaneous Localization and Mapping}
\acrodef{SNN}[SNN]{Spiking Neural Network}
\acrodef{SNR}[SNR]{Signal to Noise Ratio}
\acrodef{SOC}[SOC]{System-On-Chip}
\acrodef{SOI}[SOI]{Silicon on Insulator}
\acrodef{SP}[SP]{Separation Property}
\acrodef{SRAM}[SRAM]{Static Random Access Memory}
\acrodef{SRNN}[SRNN]{Spiking Recurrent Neural Network}
\acrodef{SRNN}[SRNN]{Spiking Recurrent Neural Networks}
\acrodef{SSC}[SSC]{Spiking Speech Command}
\acrodef{STD}[STD]{Short-Term Depression}
\acrodef{STDP}[STDP]{Spike-Timing Dependent Plasticity}
\acrodef{STP}[STP]{Short-Term Plasticity}
\acrodef{STT}[STT]{Spin-Transfer Torque}
\acrodef{STT-MRAM}[STT-MRAM]{Spin-Transfer Torque Magnetic Random Access Memory}
\acrodef{SW}[SW]{Software}
\acrodef{sWTA}[sWTA]{soft Winner-Take-All}
\acrodef{TCAM}[TCAM]{Ternary Content-Addressable Memory}
\acrodef{TFT}[TFT]{Thin Film Transistor}
\acrodef{TPU}[TPU]{Tensor Processing Unit}
\acrodef{USB}[USB]{Universal Serial Bus}
\acrodef{VHDL}[VHDL]{VHSIC Hardware Description Language}
\acrodef{VLSI}[VLSI]{Very Large Scale Integration}
\acrodef{VOR}[VOR]{Vestibulo-Ocular Reflex}
\acrodef{WCST}[WCST]{Wisconsin Card Sorting Test}
\acrodef{WL}[WL]{Word Line}
\acrodef{WTA}[WTA]{Winner-Take-All}
\acrodef{XML}[XML]{eXtensible Mark-up Language}
\acrodef{SAC}[SAC]{Soft Actor Critic}
\acrodef{PD}[PD]{Policy Distillation}
\acrodef{STE}[STE]{Straight-Through Estimator}
\acrodef{SOTA}[SOTA]{State of the Art}
\acrodef{SSM}[SSM]{State Space Model}
\acrodef{GRU}[GRU]{Gated Recurrent Unit}
\acrodef{LSTM}[LSTM]{Long Short Term Memory}
\acrodef{TCN}[TCN]{Temporal Convolutional Network}
\acrodef{FeFET}[FeFET]{Ferroelectric Field Effect Transistor}
\acrodef{JEPA}[JEPA]{Joint-Embedding Predictive Architectures}
\acrodef{RPL}[RPL]{Recurrent Predictive Learning}
\acrodef{CPC}[CPC]{Contrastive Predictive Coding}
\acrodef{RTRL}[RTRL]{Real-Time Recurrent Learning}
\acrodef{LRU}[LRU]{Linear Recurrent Unit}
\acrodef{NoC}[NoC]{Network-on-Chip}

\title{Input-dependent Neuromodulation Enhances Mixed-Signal Neuromorphic Hardware}

\author{\IEEEauthorblockN{Aashman Mehta\textsuperscript{1,2}, AbdelQader AlKilany\textsuperscript{3}, Dan F. M. Goodman\textsuperscript{3},
Giacomo Indiveri\textsuperscript{1*} and
Melika Payvand\textsuperscript{1*}}
\IEEEauthorblockA{\textsuperscript{1}Institute of Neuroinformatics, University of Zurich and ETH Zurich, Zurich, Switzerland}
\IEEEauthorblockA{\textsuperscript{2}University of California, Los Angeles, CA, USA}
\IEEEauthorblockA{\textsuperscript{3}Imperial College London, London, UK}
\thanks{We acknowledge funding support from the Swiss National Science Foundation (UNITE: TMSGI2-211461). }}

\maketitle
\begingroup
\renewcommand\thefootnote{*}
\footnotetext{Shared senior authorship}
\endgroup

\begin{abstract}

Gating mechanisms allow a fixed neural network to adapt its computation to the current input by dynamically modulating its effective parameters. In the brain, neuromodulation uses this principle to move neural circuits between dynamical regimes, and deep learning exploits input-dependent gating, e.g., in gated recurrent networks and self-attention. Input-dependent modulation has also recently been shown to improve spiking neural networks, with particularly strong benefits in small, parameter-limited ones. This makes neuromodulation especially promising for neuromorphic processors at the edge, where area, memory and energy budgets constrain network size, while circuit physics limits parameter precision. Here we demonstrate input-dependent, within-trial neuromodulation of a spiking recurrent neural network on a mixed-signal neuromorphic processor. An off-chip modulator reads the on-chip network's input and activity and adjusts its population-level gains while the chip's analog circuits compute with the physics of the silicon. Neuromodulation raises binary spoken-digit classification accuracy by $\mathbf{+17.3 \pm 10.4}$ percentage points, tested across five hardware network instantiations.

\end{abstract}

\begin{IEEEkeywords}
gating, neuromorphic hardware, spiking neural networks, mixed-signal circuits, neuromodulation
\end{IEEEkeywords}

\section{Introduction}
\label{sec:introduction}

Neuromodulation is a mechanism by which neural circuits change their computation without altering their wiring.
Concretely, neuromodulators act on existing synapses and neurons to rescale and adapt synaptic weights and neuronal gains.
For example, a single modulatory signal can multiply the strength of many synapses simultaneously, effectively reweighting the circuit's connectivity on a timescale of milliseconds to seconds, far faster than any structural change~\cite{Marder12,McCormick92}.
This is critically important because it allows a fixed anatomical circuit to express multiple, distinct dynamical regimes in response to context, arousal, or input statistics, rather than requiring separate dedicated hardware for each behavioral state~\cite{Nadim_Bucher14,Ferguson_Cardin20}.
As neuromodulation decouples the structure of a network (which is slow to change) from its effective parameters (which can be rapidly reconfigured), it greatly expands the repertoire of computations accessible to a given set of connections~\cite{HarrisWarrick89}.
Machine learning exploits a similar principle, where slowly learned parameters define a base network, while faster, input-dependent variables rescale its effective weights or gate its state updates on the timescale of a single input.
For example, in  \acp{GRU}~\cite{chung2014empirical,ha2017hypernetworks} and slow--fast weight systems~\cite{schlag2021linear}, slowly learned parameters generate fast, context-dependent rescalings of the network's effective synaptic weights.

\begin{figure}[t]
\centering
\includegraphics[width=0.9\columnwidth]{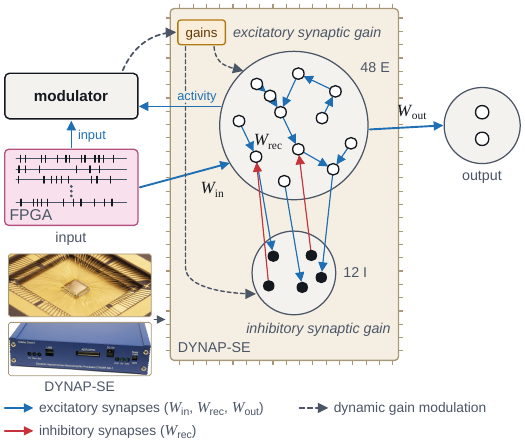}
\caption{Input-dependent gain modulation on neuromorphic hardware.
The 96 input channels are streamed into an on-chip primary network of excitatory (E) and inhibitory (I) neurons; two output neurons read the E population activity.
A modulator reads the input and the network's activity and every 40\,ms updates the excitatory and inhibitory gains of the E neurons and the gain of the readout.
The modulator and output neurons run and are trained in software, while the synaptic weights of the primary network stay fixed.}
\label{fig:arch}
\end{figure}

\citeauthor{alkilany2025neuromodulation} recently applied this principle to \acp{SNN}~\cite{alkilany2025neuromodulation}.
By using a separate modulator to dynamically adjust the parameters of a recurrent \ac{SNN}, they showed that input-dependent modulation can substantially improve performance, particularly in small networks.
These gains were not explained by added parameters alone, demonstrating that neuromodulation can increase computational capability without simply scaling network size.
Input-dependent modulation is particularly attractive in this setting, where a small number of dynamically controlled gains can reconfigure the dynamics of an existing substrate during computation.
Whether this advantage survives the constraints, variability and limited trainability of physical neuromorphic hardware, however, remains unknown.

In this work, we identify and exploit a circuit primitive in mixed-signal neuromorphic hardware whose gain can be dynamically modulated to implement fast, input-dependent gating.
We implement the recurrent network as a fixed physical reservoir, while a separate modulator adjusts a small number of shared population-level gains, thereby reconfiguring the network's computation without changing its underlying connectivity.
Because gradients cannot be propagated through the analog circuits, the modulator is trained from the reservoir's measured response to controlled changes in these gain parameters. Fig.~\ref{fig:arch} illustrates the resulting closed-loop system, in which the modulator observes the input and evolving network state and continuously adapts the physical network while it computes.

We validate the system across five physical reservoirs, showing that input-dependent modulation improves five/nine spoken-digit classification from 71.4\,\% to 88.6\,\%, with gains in every reservoir.
Controls separate the effects of additional readout training, static gain optimization and input dependence.
These results demonstrate that dynamic gating can increase the computational capability of a heterogeneous analog neuromorphic computing substrate, affected by device mismatch and variability, providing a route to richer computation without having to scale the size of the underlying network.

\section{Methods}
\label{sec:methods}

The chip used in this work is the scalable Dynamic Neuromorphic Asynchronous Processor (DYNAP-SE), a multi-core neuromorphic processor with spiking silicon neurons, dynamic synapses and asynchronous address-event routing circuits~\cite{Moradi_etal18}.

\subsection{The DYNAP-SE chip}
\label{sec:dynap-se}

\begin{figure}
  \centering
  \includegraphics[width=0.3\textwidth]{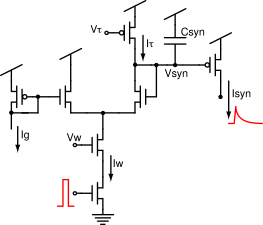}
  \caption{DPI synapse circuit schematics, highlighting the origin of the gain modulation current parameter $I_g$, the synaptic weight $I_w$, and the current determining the circuit's time constant $I_{\tau}$.}
  \label{fig:dpi}
\end{figure}

DYNAP-SE was fabricated in a 0.18\,$\mu$m 1P6M CMOS technology and comprises four cores of 256 neurons each.
Each neuron has 64 dynamic synapses, for a total of 1024 neurons and 64k synapses.
The neuron circuits implement the adaptive exponential integrate-and-fire (AdExp-I\&F) model using sub-threshold analog circuits, while synaptic dynamics are implemented using Differential Pair Integrator (DPI) log-domain filters~\cite{Bartolozzi_etal06}.
The DPI synapses approximate linear first-order dynamics~\cite{Bartolozzi_Indiveri07a}.
For each synapse type, the DPI circuit provides a synaptic weight term and a separate synaptic gain scaling term (see Fig.~\ref{fig:dpi}).
As derived in~\cite{Chicca_etal14}, to a first-order approximation, the DPI output synaptic current $I_{syn}$ is governed by the following equation:
\[
  \tau \frac{d}{dt}I_{syn} + I_{syn} = I_w \frac{I_g}{I_{\tau}}, \quad \tau = \frac{C_{syn} U_T}{\kappa I_{\tau}}
\]
where $\tau$ is the synapse time constant, $I_w$ represents the synaptic weight, $I_g$ represents a gain modulation factor, $C_{syn}$ is the synapse capacitance, $U_T$ the thermal voltage, and $\kappa$ the subthreshold slope factor~\cite{Liu_etal02a}.

The DYNAP-SE chip supports multiple synaptic dynamics; in this work we use DPI circuits configured to emulate fast excitatory (AMPA-like) and shunting inhibitory ($\mathrm{GABA_A}$-like) synapses.
The corresponding weight, gain, and time constant parameters of each synapse type are shared by all neurons within a core.
However, as device mismatch introduces neuron-to-neuron and synapse-to-synapse variability, with coefficients of variation of up to 0.2, synapses and neurons in the network have a degree of variability that can be helpful in shaping computation~\cite{Dahmen_etal26,Zendrikov_etal23}.
The shared synaptic gains are the parameters that we modulate dynamically in this work.

To account for the DPI bias range restrictions, we restrict modulation of the $I_g$ terms to a band $[g_0/b,\,g_0b]$ around each calibrated baseline gain $g_0$, with $b=1.3$.

\subsection{Task setup}
\label{sec:validation}

We evaluate the networks on binary classification of the digits five and nine from the Spiking Heidelberg Digits (SHD) dataset~\cite{Cramer_etal22}.
We use a binary task to keep the problem within reach of the small networks that can be implemented on chip while retaining sufficient difficulty to measure the effect of modulation.
A sweep over all binary SHD pairs selected five/nine, for which small recurrent networks perform appreciably above chance and outperform feed-forward networks.

The 700 SHD input channels are compressed to 96 by summing groups of 7--8 contiguous channels, allowing the input to be mapped within the 64-synapse fan-in constraint of each DYNAP-SE neuron.
For each network, we use five speaker-disjoint training/validation splits, each holding out a different pair of the 10 training speakers for validation.
Testing uses the 213 five/nine trials of the standard SHD test split, which includes two speakers not present during training.

\subsection{On-chip primary network}
\label{sec:primary-network}

Because gradients cannot be propagated through the mixed-signal dynamics of DYNAP-SE, we use the on-chip recurrent network as a fixed physical reservoir and train only an off-chip readout.

The reservoir consists of 48 excitatory (E) and 12 inhibitory (I) neurons placed on separate DYNAP-SE cores, giving the two populations independent shared parameters and enforcing Dale's law~\cite{Strata_Harvey99}.
The membrane time constant is set to $\tau_\mathrm{mem}=40$\,ms, while the fast excitatory and shunting inhibitory synapses use $\tau_\mathrm{AMPA}=\tau_{\mathrm{GABA_A}}=10$\,ms.

As shown in Fig.~\ref{fig:arch}, both $W_\mathrm{in}$ and $W_\mathrm{rec}$ are fixed and randomly drawn, with a fixed number of connections per neuron.
$W_\mathrm{in}$ gives each E neuron 6 input channels.
In $W_\mathrm{rec}$, each E neuron receives excitatory synapses from 20 other E neurons and inhibitory synapses from all 12 I neurons, and in turn projects to 6 of the 12 I neurons.
The activity of the E population is read out through $W_\mathrm{out}$ by two off-chip spiking output units, one per class, and $W_\mathrm{out}$ is trained in software with a cross-entropy loss $L$ on their spike counts, whose argmax gives the predicted class.

This setup lets us test whether input-dependent modulation of a small number of population gains can increase the computational capability of an otherwise fixed physical reservoir.

\begin{figure*}
  \centering
  \begin{subfigure}[t]{0.32\textwidth}
    \includegraphics[width=\textwidth]{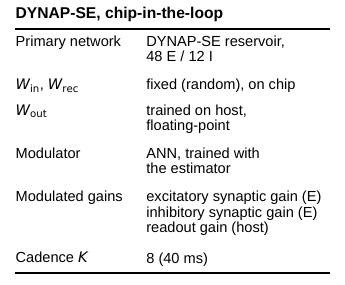}
    \caption{}
    \label{fig:chipsetup}
  \end{subfigure}
  \hfill
  \begin{subfigure}[t]{0.32\textwidth}
    \includegraphics[width=\textwidth]{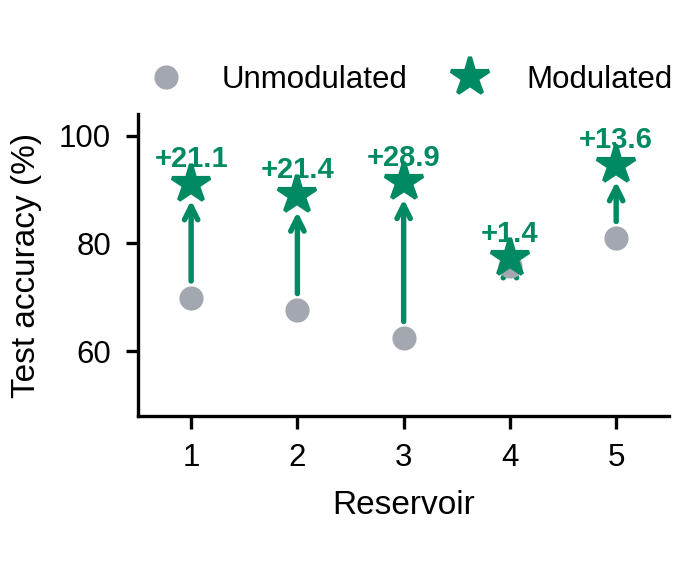}
    \caption{}
    \label{fig:chipa}
  \end{subfigure}
  \hfill
  \begin{subfigure}[t]{0.32\textwidth}
    \includegraphics[width=\textwidth]{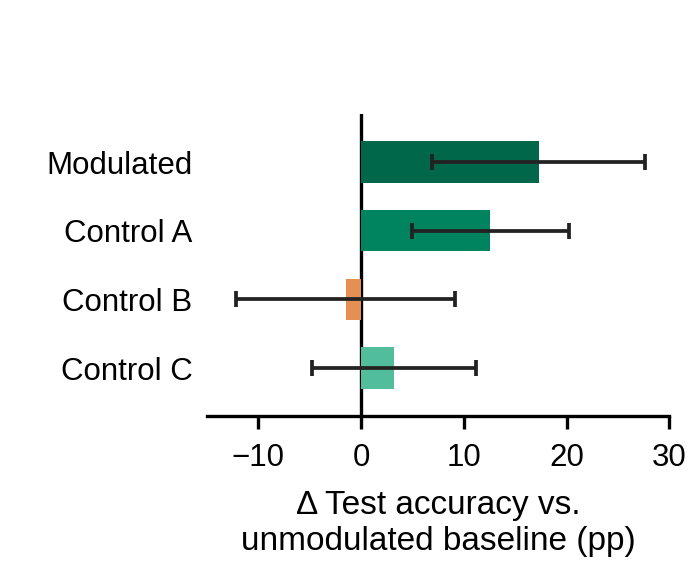}
    \caption{}
    \label{fig:chipb}
  \end{subfigure}
  \caption{Chip-in-the-loop modulation on DYNAP-SE (5 RNNs; excitatory and inhibitory gains of the E neurons on chip and somatic gain for off-chip readout neurons rewritten every 40\,ms; 213 test trials); (\subref{fig:chipsetup}) What is fixed, trained and modulated; (\subref{fig:chipa}) Test accuracy of each RNN without and with modulation, the modulator trained with the activity-gated estimated gradient; (\subref{fig:chipb}) Change in test accuracy over the unmodulated baseline for the modulated network and Controls~A--C, mean~$\pm$~SD over RNNs.}
  \label{fig:chip}
\end{figure*}

\subsection{Modulator and gain updates}
\label{sec:modulators}

The modulator closes the feedback loop shown in Fig.~\ref{fig:arch}.
It is implemented off chip as a multilayer perceptron with one hidden layer of 128 ReLU units and three tanh output units, one for each modulated gain.

The modulator observes the 96 input channels together with the activity of the 48 excitatory neurons, 12 inhibitory neurons, and two output units.
These signals are averaged over 40\,ms windows and provided together with the current gain values.
At the end of each window, the modulator produces three signed outputs $\delta_n \in [-1,1]$ that independently update the excitatory and inhibitory synaptic gains onto the excitatory population and, on the host, the gain of the readout.

The modulator updates the logarithm of each gain.
For each gain, an internal variable $a_n$, initialized as $a_0=\ln g_0$, integrates the modulator output,
\begin{equation}
    a_n = a_{n-1} + s\,\delta_n,
    \qquad
    g_n = \exp(a_n),
\end{equation}
where $g_0$ is the calibrated baseline gain.
The resulting gain is soft-clipped to the hardware operating range $[g_0/b,\,g_0b]$, with $b=1.3$.
The step size $s$ is chosen such that the full modulation range can be reached within a single trial.

The updated gains are written to DYNAP-SE at runtime every 40\,ms without resetting or interrupting the continuous-time dynamics of the reservoir.

\subsection{Chip-in-the-loop training}
\label{sec:chip-training}

Because gradients cannot be propagated through the analog dynamics of DYNAP-SE, we train the modulator using an estimate of the reservoir's response to changes in its gain parameters.
Before training, reservoir activity is recorded for all training and validation trials over a $4\times4$ grid spanning the allowed range of the two modulated gains, with two repeats per setting.
For each trial and excitatory neuron, a quadratic function of the logarithmic gains is fitted by least squares to the firing rate averaged over the trial.
This provides an estimate of how each neuron's activity changes with each gain.

During training, backpropagation through the off-chip readout provides $\partial L/\partial r_{i,w}$ for the firing rate $r_{i,w}$ of each excitatory neuron in each modulation window $w$.
Combining this with the derivative of the fitted response gives an estimated gradient for gain $b$ in modulation window $w$,
\begin{equation}
\hat{g}_{b,w}
=
[A_w \geq \tau_g]
\sum_{i=1}^{48}
\frac{\partial L}{\partial r_{i,w}}
\frac{\partial \hat{r}_i}{\partial \theta_b}
\bigg|_{\theta_w},
\end{equation}
where $\hat{r}_i$ is the fitted firing rate, $\theta_b$ is the logarithm of gain $b$, $\theta_w$ are the log-gains the modulator sets at the end of window $w$, and $A_w$ is the input activity in the window.
For windows with insufficient input activity, the estimated gradient is set to zero because the fitted response is unreliable.

We first validated this gradient estimator in a DYNAP-aware simulation and found performance comparable to backpropagation through the reservoir.
The same simulations confirmed that gain modulation remains effective with a 40\,ms update interval, which we therefore use for the hardware experiments.

The recording grid grows rapidly with the number of modulated parameters, so on chip we restrict modulation to the excitatory and inhibitory synaptic gains onto the E population.
The excitatory synaptic gain is used rather than the somatic gain because it produced an approximately $5\times$ larger measured effect on the chip.

For each reservoir, $W_\mathrm{out}$ is first trained from activity recorded at the baseline gains.
The readout and modulator are then trained jointly for 20 epochs with DYNAP-SE in the forward loop, using the estimated gain gradients to train the modulator.
The epoch with the best validation accuracy is retained.

\subsection{Controls and statistics}
\label{sec:controls}



Three controls test which aspects of gain modulation account for the performance improvement.
Control~A retrains the readout without modulation, testing whether additional readout training alone explains the improvement.
For Controls~B and~C, the trained readout and modulator are kept fixed and only the applied gain trajectories are changed.
Control~B replaces the dynamic gains by one constant setting for all trials, equal to their mean over the validation trials, testing whether modulation simply finds a better fixed set of gains.
Control~C applies to each trial the gain trajectory generated by another trial, testing whether modulation must depend on the current input.
The controls are summarized in Table~\ref{tab:controls}.

\begin{table}[t]
\centering
\caption{Controls used in the hardware experiments.}
\label{tab:controls}
\small
\renewcommand{\arraystretch}{1.15}
\begin{tabular}{@{}c>{\raggedright\arraybackslash}p{0.46\columnwidth}>{\raggedright\arraybackslash}p{0.42\columnwidth}@{}}
\hline\noalign{\vskip 2pt}
  & Modification & Tests whether improvement requires \\[2pt]
\hline\noalign{\vskip 2pt}
A & Retrain readout only (fresh optimizer, same epochs) & Modulation beyond extra readout training \\[2pt]
B & Gains held at the mean of the modulated validation run & Dynamic gains rather than a better operating point \\[2pt]
C & Another trial's gain trajectory played through the reservoir & Dependence of the gains on the current input \\[2pt]
\hline
\end{tabular}
\end{table}

Results are compared across the five physical reservoirs using two-sided paired $t$-tests.
$p$-values are corrected for multiple comparisons ($q$-values) using the Benjamini--Hochberg procedure, and values are reported as mean~$\pm$~SD across reservoirs.

\section{Experimental Results}

We evaluate input-dependent gain modulation on five physical DYNAP-SE reservoirs using the chip-in-the-loop training procedure described in Sec.~\ref{sec:chip-training}.
The excitatory and inhibitory synaptic gains onto the E population, and the readout gain on the host, are updated every 40\,ms while the reservoir processes the input.

Modulation increases five/nine test accuracy from 71.4\,\% to 88.6\,\%, corresponding to an improvement of $+17.3 \pm 10.4$ percentage points (pp) across the five reservoirs ($p = 0.021$, $q = 0.082$).
The improvement is positive for every physical reservoir (Fig.~\ref{fig:chipa}).
Holding the readout gain at a constant, its validation mean, lowers accuracy by only $0.4 \pm 0.8$\,pp ($p = 0.34$, $q = 0.45$): the improvement comes from modulating the on-chip gains.

The controls in Fig.~\ref{fig:chipb} separate this improvement from alternative explanations.
Control~A, which only retrains the readout, reaches 83.9\,\%, $4.7 \pm 7.7$\,pp below the modulated system ($p = 0.24$, $q = 0.30$).
Control~B reduces accuracy by $18.8 \pm 13.1$\,pp ($p = 0.033$, $q = 0.082$), compared to the modulated system, showing that a better static operating point does not explain the gain.
Control~C reduces accuracy by $14.1 \pm 11.4$\,pp ($p = 0.051$, $q = 0.085$), testing the importance of input dependence for gain modulation.


Together, these results show that dynamically modulating a small number of population-level gains can substantially improve the computational capability of a fixed physical reservoir.




\section{Conclusion}

We demonstrate input-dependent, within-trial gain modulation on DYNAP-SE, a mixed-signal neuromorphic processor.
An off-chip ANN modulator dynamically adjusts the excitatory and inhibitory synaptic gains of a fixed 60-neuron physical reservoir using gradients estimated from measured chip responses.
Across five physical reservoirs, modulation improves five/nine classification accuracy from 71.4\,\% to 88.6\,\%, corresponding to an improvement of $+17.3 \pm 10.4$ percentage points.
The control experiments further separate this improvement from readout retraining and static gain optimization, and test the role of input dependence.

These results show that a small number of dynamically controlled population-level gains can increase the computational capability of an otherwise fixed and heterogeneous analog substrate.
Rather than changing the underlying connectivity, the modulator reconfigures the effective computation of the reservoir on the timescale of the input.

The system is not yet fully hardware-native, as the modulator and readout run on the host, and the present experiments consider one binary task and two modulated on-chip gains.
A next step is to move these functions closer to the neuromorphic substrate and ultimately realize modulation directly through on-chip circuitry.
Such systems could provide a path toward neuromorphic processors whose physical dynamics can follow the input and computational context.

\section*{Acknowledgment}
We thank Mirco Tincani for help with the DYNAP-SE setup and Filippo Moro for helpful discussions on the experiments. This project benefited from discussions and collaborations at the 2026 Capocaccia Neuromorphic Intelligence workshop.
The authors used Claude (Anthropic) for language editing and grammar of Sections I--IV. Code used for data processing, sweep scripts and drawing the Fig.~1 schematic was partly AI-assisted. All content was reviewed and verified by the authors.

\printbibliography

@Article{Bartolozzi_Indiveri07a,
author		= {Chiara Bartolozzi and Giacomo Indiveri},
title		= {Synaptic dynamics in analog {VLSI}},
journal		= {Neural Computation},
year		= {2007},
month		= oct,
volume		= {19},
number		= {10},
pages		= {2581--2603},
doi		= {10.1162/neco.2007.19.10.2581}
}

@InProceedings{Bartolozzi_etal06,
author		= {Chiara Bartolozzi and Srinjoy Mitra and Giacomo Indiveri},
title		= {An ultra low power current--mode filter for neuromorphic
		  systems and biomedical signal processing},
booktitle	= {{IEEE} Biomedical Circuits and Systems Conference
		  ({BioCAS})},
year		= {2006},
pages		= {130--133},
organization	= {IEEE},
doi		= {10.1109/BIOCAS.2006.4600325}
}

@Misc{CAVIAR,
key		= {CAVIAR},
title		= {Convolution Address-Event-Representation ({AER}) Vision
		  Architecture for Real-Time ({CAVIAR})},
howpublished	= {IST-2001-34124 EU Grant},
year		= {2002}
}

@Article{Chicca_etal14,
author		= {Elisabetta Chicca and Fabio Stefanini and Chiara
		  Bartolozzi and Giacomo Indiveri},
title		= {Neuromorphic electronic circuits for building autonomous
		  cognitive systems},
journal		= {Proceedings of the {IEEE}},
year		= {2014},
month		= sep,
volume		= {102},
number		= {9},
pages		= {1367--1388},
issn		= {0018-9219},
doi		= {10.1109/JPROC.2014.2313954}
}

@Misc{DVS,
key		= {DVS},
title		= {Product of inilabs: Dynamic Vision Sensor},
howpublished	= {http://www.inilabs.com/products/dynamic-vision-sensors},
year		= {2009}
}

@Article{Dahmen_etal26,
author		= {Dahmen, David and Hutt, Axel and Indiveri, Giacomo and
		  Kennedy, Ann and Lefebvre, Jeremie and Mazzucato, Luca and
		  Motter, Adilson E. and Narayanan, Rishikesh and Payvand,
		  Melika and Planert, Henrike and Gast, Richard},
title		= {How heterogeneity shapes dynamics and computation in the
		  brain},
journal		= {Neuron},
year		= {2026},
month		= mar,
publisher	= {Elsevier BV},
issn		= {0896-6273},
doi		= {10.1016/j.neuron.2025.11.023},
url		= {http://dx.doi.org/10.1016/j.neuron.2025.11.023}
}

@Book{Liu_etal02a,
author		= {Shih-Chii Liu and J. Kramer and Giacomo Indiveri and Tobi
		  Delbruck and Rodney J. Douglas},
title		= {Analog {VLSI}:~Circuits and Principles},
year		= {2002},
publisher	= {MIT Press}
}

@Article{Marder12,
author		= {Marder, Eve},
title		= {Neuromodulation of {Neuronal} {Circuits}: {Back} to the
		  {Future}},
journal		= {Neuron},
year		= {2012},
month		= oct,
volume		= {76},
number		= {1},
pages		= {1--11},
issn		= {0896-6273},
doi		= {10.1016/j.neuron.2012.09.010},
shorttitle	= {Neuromodulation of {Neuronal} {Circuits}},
language	= {en}
}

@Article{Moradi_etal18,
author		= {Moradi, Saber and Qiao, Ning and Stefanini, Fabio and
		  Indiveri, Giacomo},
title		= {A Scalable Multicore Architecture With Heterogeneous
		  Memory Structures for Dynamic Neuromorphic Asynchronous
		  Processors ({DYNAPs})},
journal		= {{IEEE} Transactions on Biomedical Circuits and Systems},
year		= {2018},
month		= feb,
volume		= {12},
number		= {1},
pages		= {106--122},
doi		= {10.1109/TBCAS.2017.2759700}
}

@Article{Strata_Harvey99,
author		= {Piergiorgio Strata and Robin Harvey},
title		= {Dale's principle},
journal		= {Brain Research Bulletin},
year		= {1999},
volume		= {50},
number		= {5},
pages		= {349--350},
issn		= {0361-9230},
doi		= {10.1016/S0361-9230(99)00100-8},
url		= {https://www.sciencedirect.com/science/article/pii/S0361923099001008}
}

@Article{Zendrikov_etal23,
author		= {Zendrikov, Dmitrii and Solinas, Sergio and Indiveri,
		  Giacomo},
title		= {Brain-inspired methods for achieving robust computation in
		  heterogeneous mixed-signal neuromorphic processing
		  systems},
journal		= {Neuromorphic Computing and Engineering},
year		= {2023},
month		= jul,
volume		= {3},
number		= {3},
pages		= {034002},
publisher	= {IOP Publishing},
issn		= {2634-4386},
doi		= {10.1088/2634-4386/ace64c},
url		= {http://dx.doi.org/10.1088/2634-4386/ace64c}
}

@inproceedings{schlag2021linear,
  title={Linear transformers are secretly fast weight programmers},
  author={Schlag, Imanol and Irie, Kazuki and Schmidhuber, J{\"u}rgen},
  booktitle={International conference on machine learning},
  pages={9355--9366},
  year={2021},
  organization={PMLR}
}

@article{chung2014empirical,
  title={Empirical Evaluation of Gated Recurrent Neural Networks on Sequence Modeling},
  author={Chung, Junyoung and Gulcehre, Caglar and Cho, Kyunghyun and Bengio, Yoshua},
  journal={arXiv preprint arXiv:1412.3555},
  year={2014}
}

@inproceedings{ha2017hypernetworks,
  title={HyperNetworks},
  author={Ha, David and Dai, Andrew M and Le, Quoc V},
  booktitle={International Conference on Learning Representations (ICLR)},
  year={2017},
  url={https://openreview.net/forum?id=rkpACe1lx}
}

@article{alkilany2025neuromodulation,
  title={Neuromodulation enhances dynamic sensory processing in spiking neural network models},
  author={AlKilany, AbdalQader and Goodman, Dan FM},
  journal={bioRxiv},
  pages={2025--07},
  year={2025},
  publisher={Cold Spring Harbor Laboratory}
}

@article{edvac,
  title={First draft of a report on the EDVAC},
  author={von Neumann, John},
  year={1945},
  note={Moore School of Electrical Engineering, University of Pennsylvania}
}

@article{Nadim_Bucher14,
  author  = {Nadim, Farzan and Bucher, Dirk},
  title   = {Neuromodulation of neurons and synapses},
  journal = {Current Opinion in Neurobiology},
  volume  = {29},
  pages   = {48--56},
  year    = {2014},
  doi     = {10.1016/j.conb.2014.05.003}
}

@article{Ferguson_Cardin20,
  author  = {Ferguson, Katie A. and Cardin, Jessica A.},
  title   = {Mechanisms underlying gain modulation in the cortex},
  journal = {Nature Reviews Neuroscience},
  volume  = {21},
  number  = {2},
  pages   = {80--92},
  year    = {2020},
  doi     = {10.1038/s41583-019-0253-y}
}

@article{McCormick92,
  author  = {McCormick, David A.},
  title   = {Neurotransmitter actions in the thalamus and cerebral cortex and their role in neuromodulation of thalamocortical activity},
  journal = {Progress in Neurobiology},
  volume  = {39},
  number  = {4},
  pages   = {337--388},
  year    = {1992},
  doi     = {10.1016/0301-0082(92)90012-4}
}

@article{Cramer_etal22,
  author  = {Cramer, Benjamin and Stradmann, Yannik and Schemmel, Johannes and Zenke, Friedemann},
  title   = {The {Heidelberg} Spiking Data Sets for the Systematic Evaluation of Spiking Neural Networks},
  journal = {{IEEE} Transactions on Neural Networks and Learning Systems},
  volume  = {33},
  number  = {7},
  pages   = {2744--2757},
  year    = {2022},
  doi     = {10.1109/TNNLS.2020.3044364}
}

@inproceedings{HarrisWarrick89,
  author    = {Harris-Warrick, Ronald M.},
  title     = {Mechanisms for Neuromodulation of Biological Neural Networks},
  booktitle = {Advances in Neural Information Processing Systems 2 (NIPS 1989)},
  volume    = {2},
  pages     = {18--27},
  year      = {1989}
}

\end{document}